\documentclass[letterpaper, 10 pt, conference]{ieeeconf}  

\IEEEoverridecommandlockouts                              

\usepackage{graphicx} 
\usepackage{amsmath} 
\usepackage{float}
\usepackage{caption}
\usepackage{url}
\usepackage{booktabs}
\usepackage{makecell}

\title{\LARGE \bf
SAM2Grasp: Resolve Multi-modal Grasping via Prompt-conditioned Temporal Action Prediction
}

\author{
Shengkai Wu$^1$, 
Jinrong Yang, 
Wenqiu Luo, 
Linfeng Gao, 
Chaohui Shang, 
Meiyu Zhi, 
Mingshan Sun,
\\
Fangping Yang, 
Liangliang Ren, 
and Yong Zhao$^{*}$%
\thanks{$^{*}$Corresponding author: zhaoyong11933@cvte.com; $^{1}$First author: wushengkai@cvte.com. All authors are with CVTE, Guangzhou, China.}
}

\begin{document}

\maketitle
\thispagestyle{empty}
\pagestyle{empty}

\begin{abstract}

Imitation learning for robotic grasping is often plagued by the multimodal problem: when a scene contains multiple valid targets, demonstrations of grasping different objects create conflicting training signals. Standard imitation learning policies fail by averaging these distinct actions into a single, invalid action. In this paper, we introduce SAM2Grasp, a novel framework that resolves this issue by reformulating the task as a uni-modal, prompt-conditioned prediction problem. Our method leverages the frozen SAM2 model to use its powerful visual temporal tracking capability and introduces a lightweight, trainable action head that operates in parallel with its native segmentation head. This design allows for training only the small action head on pre-computed temporal-visual features from SAM2. During inference, an initial prompt, such as a bounding box provided by an upstream object detection model, designates the specific object to be grasped. This prompt conditions the action head to predict a unique, unambiguous grasp trajectory for that object alone. In all subsequent video frames, SAM2's built-in temporal tracking capability automatically maintains stable tracking of the selected object, enabling our model to continuously predict the grasp trajectory from the video stream without further external guidance. This temporal-prompted approach effectively eliminates ambiguity from the visuomotor policy. We demonstrate through extensive experiments that SAM2Grasp achieves state-of-the-art performance in cluttered, multi-object grasping tasks.

\end{abstract}

\section{INTRODUCTION}

\begin{figure}[htbp]
    \centering
    \includegraphics[width=0.9\columnwidth]{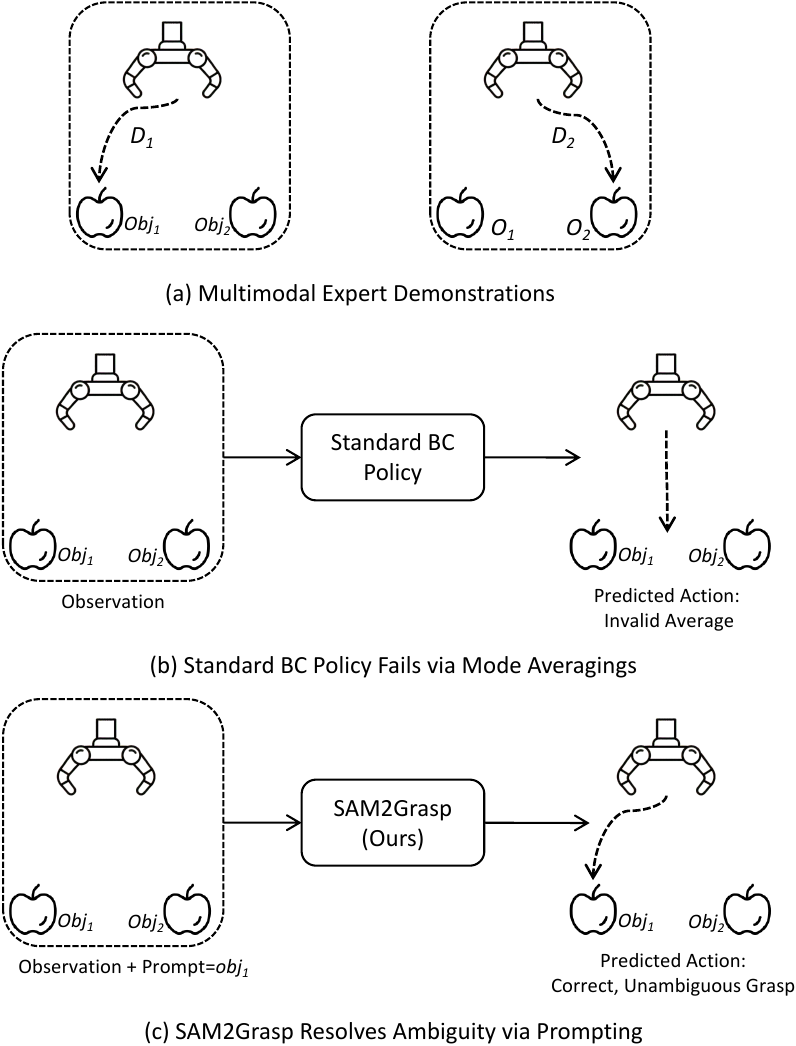}
    \caption{
        \textbf{The Multi-modality Problem in Imitation Learning and Our Approach.} 
        (a) Given a single observation with multiple valid objects, expert demonstrations can be multi-modal (e.g., grasping the left or right object). 
        (b) A standard Behavioral Cloning (BC) policy trained on this data fails by "mode averaging," predicting a physically nonsensical action that targets the empty space between objects. 
        (c) Our method, SAM2Grasp, resolves this ambiguity by taking an additional prompt that specifies the target. This transforms the multi-modal problem into a uni-modal one, enabling the policy to generate a correct and unambiguous action.
    }
    \label{fig:teaser}
\end{figure}

Learning robotic policies from demonstration, often formulated as a supervised regression task of mapping observations to actions, is a compelling paradigm for acquiring complex visuomotor skills. In practice, however, predicting robot actions presents unique challenges, chief among them being the inherent multi-modality of action distributions. This problem is ubiquitous in grasping, where, for example, a single observation containing several objects affords multiple, equally valid actions, as depicted in Figure~\ref{fig:teaser}(a). When trained on such data, a standard behavioral cloning policy fails by predicting the "average" of distinct expert actions—a physically meaningless command that grasps empty space, as illustrated in Figure~\ref{fig:teaser}(b).

Prior work has largely focused on tackling this challenge by engineering more expressive policy representations capable of modeling this complexity. These strategies include explicitly modeling a multi-modal output with mixture of Gaussians \cite{c2}, or switching to implicit policy representations via generative models like CVAE \cite{cvae}  and Diffusion Models\cite{ddpm} to better capture the diverse distribution of expert behaviors. While effective, these methods inherently ask the policy to manage the ambiguity, adding significant complexity to the learning problem.

This paper presents a fundamentally different strategy: we resolve ambiguity at the perception stage rather than modeling it at the action stage. As shown in Figure~\ref{fig:teaser}(c), we introduce SAM2Grasp, a framework that conditions the policy on a single, explicit target via a prompt. This simple but powerful idea transforms the intractable multi-modal problem into a well-defined, uni-modal one. We realize this approach by leveraging a powerful, pre-trained foundation model, the Segment Anything Model 2 (SAM2)~\cite{sam2}. Its built-in capabilities for promptable, temporal-aware tracking provide the perfect tool to eliminate ambiguity upfront, thus dramatically simplifying the downstream policy learning task.

Our architecture, which integrates a frozen SAM2 backbone with a lightweight ACT policy head, is intentionally designed for efficiency. This choice unlocks a remarkably efficient training paradigm. Instead of costly end-to-end training, we perform a single offline pass to extract and cache rich temporal-visual features from SAM2 across our entire demonstration dataset. Subsequently, only the action head is trained to map these powerful, pre-computed features to grasp trajectories. This approach avoids the immense computational cost and data requirements of fine-tuning the large foundation model.

At inference time, SAM2Grasp operates as a robust, closed-loop policy. An initial prompt such as a bounding box from an upstream module specifies the target object. The model then leverages SAM2's temporal tracking capabilities to autonomously follow the designated object across subsequent video frames, continuously refining its trajectory without requiring further prompts. This "temporal-prompted" execution transforms the policy's task from a multi-choice problem into a deterministic, conditioned execution, ensuring robust performance even under visual occlusion.

Our contributions are:

\begin{itemize}
    \item We introduce SAM2Grasp, a novel framework that resolves object-level multimodality in robotic grasping by reformulating it as a prompt-conditioned, uni-modal problem.
    \item We propose an efficient architecture that deeply integrates a frozen, temporal-aware foundation model (SAM2) with an ACT policy, demonstrating a powerful paradigm of decoupling perception from control.
    \item We present extensive experiments showing that SAM2Grasp not only achieves state-of-the-art performance but also exhibits robustness to severe visual occlusions, dramatically outperforming SOTA baselines.
\end{itemize}

\section{RELATED WORK}

\subsection{Imitation Learning for Robotic Grasping}

Imitation learning (IL) and its most direct form, Behavioral Cloning (BC)~\cite{bc}, cast robot skill acquisition as a supervised learning problem mapping observations to actions. Subsequent research has significantly advanced this paradigm. A major thrust has been to improve policy architectures to better incorporate history, evolving from simple feed-forward networks to Transformer-based models that capture long-term dependencies in sensorimotor data~\cite{rt1}. Another direction focuses on enhancing the versatility of IL, leveraging language for multi-task learning~\cite{rt2,pi0,pi05,openvla,fast,smolvla} or scaling with massive datasets to achieve impressive generalization~\cite{rt1, rtx,graspvla,gen0}. In addition, high-quality demonstration data is important for IL training and some works have explored the rules to collect high-quality data, such as HD-Space~\cite{hd} and ADC~\cite{adc}. However, a fundamental challenge persists: when trained on expert data containing multiple valid but distinct behaviors for the same observation (e.g., grasping different objects), these end-to-end policies suffer from the catastrophic mode-averaging problem. Our work directly targets this critical limitation within the end-to-end learning paradigm.

\subsection{Addressing Multi-modality in Imitation Learning}

The fundamental multi-modality problem in IL has inspired two distinct strategic directions. The first focuses on modeling ambiguity in action distribution. This is typically achieved with generative models, ranging from Conditional Variational Autoencoders (CVAEs)~\cite{cvae}, as used in the original Action Chunking Transformer (ACT)~\cite{act}, to more powerful Diffusion Models~\cite{ddpm} like in Diffusion Policy~\cite{DP}. A second, emerging strategy resolves ambiguity at the policy input by providing an explicit target condition. This is often done by augmenting the policy's observation with target-specific information, such as a target's mask ~\cite{Dexgraspvla} or a cropped image of the target object~\cite{vip}. While this visual cueing is more effective, it still requires a standard vision backbone to learn the difficult task of correlating this cue with the main visual stream from scratch. Our work, SAM2Grasp, advances this second strategy to a new level of integration and efficiency. Instead of merely concatenating a condition to the input, we leverage the built-in promptable and temporal-aware capabilities of the frozen SAM2 model, which allows us to extract clean, object-centric features that are already filtered from background clutter before they reach the policy. As our experiments demonstrate, this prompt-driven perception is fundamentally more robust than simple input conditioning, as it offloads the entire complex task of conditioned perception and tracking to a specialized, pre-trained expert.

\subsection{Foundation Models for Robotic Manipulation}

The rise of large-scale, pre-trained foundation models~\cite{depthanythingv2,depthpro,sam2} has created a paradigm shift in robotics. Vision-Language Models (VLMs) \cite{clip,palm-e, pali,siglip} have enabled robots to follow natural language instructions~\cite{rt2,pi0}. In addition, the Segment Anything Model (SAM)~\cite{sam} and its successor SAM2~\cite{sam2} introduced promptable segmentation, providing a powerful tool for zero-shot object perception, which is important for robotic manipulation. For example, SAM2Act \cite{sam2act} introduced SAM2 into a 3D-based robotic policy for high precision and generalizability. Theia\cite{theia} distills diverse vision foundation models such as SAM \cite{sam}, Depth-Anything \cite{depthanyting} for robot learning. Our work distinguishes them by deeply integrating a temporal-aware foundation model not as a mere perception module, but as a core, frozen component of the visuomotor policy itself. Instead of just using its final segmentation mask, we harness its rich internal features and built-in temporal tracking capabilities to directly drive action prediction. This deep integration allows us to create a highly efficient yet powerful policy that inherits the immense knowledge of the foundation model without the need for costly fine-tuning.

\section{Method}

\begin{figure*}[!htbp]  
    \centering
    \includegraphics[width=1.0\textwidth]{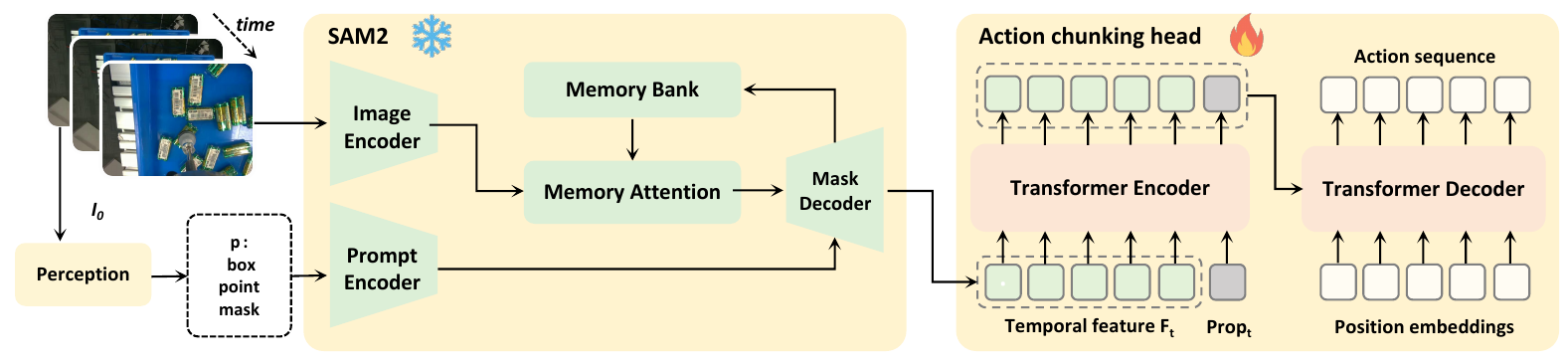}  
    \caption{
        \textbf{The SAM2Grasp Architecture.} 
Our framework uses a prompt to guide a frozen SAM2 model in extracting object-centric features, which are then fed into a trainable ACT policy head. At $t=0$, an external prompt $p$ is required. For $t>0$, SAM2's internal temporal memory handles object tracking autonomously. This design resolves object-level multimodality at the perception stage.
    }
    \label{fig:architecture}
\end{figure*}

Our goal is to develop a robust and efficient imitation learning framework for robotic grasping that resolves the fundamental multi-modality problem when grasping multiple objects. Our key insight is to reframe the task: instead of learning a complex multi-modal policy, we learn a simple, uni-modal policy conditioned on a target-object prompt. We achieve this by building upon a frozen, pre-trained temporal-aware foundation model, SAM2 \cite{sam2}. This section details our problem formulation, the SAM2Grasp architecture, our highly efficient training paradigm, and the inference-time execution pipeline.

\subsection{Problem Formulation} 

Standard Behavioral Cloning (BC) aims to learn a policy $\pi$ that minimizes a loss function between its predicted action $a_t$ and an expert's action $a_t^*$ from a demonstration dataset $\mathcal{D} = {(s_t, a_t^*)}$. However, this formulation faces significant challenges in robotic grasping due to the inherent multi-modality of expert data.

The multi-modality problem in robotic grasping manifests in at least two distinct ways:

\begin{itemize}
    \item Object-level multimodality: Given multiple graspable objects, which one should be the target? Demonstrations where experts grasp different objects from the same initial scene create conflicting data pairs $(s_t, a_{t,1}^*)$ and $(s_t, a_{t,2}^*)$, where $a_{t,1}^* \neq a_{t,2}^*$.
    \item Action-level multimodality: For a single target object, how should it be grasped? An expert may demonstrate various valid grasp poses (e.g., top-down grasp, side grasp), creating a multi-modal distribution even for a single target.
\end{itemize}

While both are valid challenges, the object-level multimodality is particularly detrimental for standard BC. Training a deterministic policy on such data leads to mode averaging, often resulting in a physically nonsensical trajectory that targets the empty space between objects. Our work explicitly addresses this object-level multimodality.

We reformulate the problem by introducing an object-specifying prompt, $p$, as an additional condition for the policy. The policy's task becomes learning the mapping $\pi(a_t | s_t, p)$. The prompt $p$ (e.g., a bounding box provided by an object detection model) uniquely identifies a target object $o_i$ within the scene. For any given pair of $(s_t, p)$, the corresponding expert action $a_t^*$ is now assumed to be uni-modal at the object-level, as the ambiguity of which object to grasp has been resolved. Our objective is to learn this simpler, conditioned policy.

\subsection{SAM2Grasp Architecture}

As illustrated in Figure 2, our SAM2Grasp architecture is designed for modularity and efficiency. It consists of two primary components: a frozen, pre-trained SAM2 \cite{sam2} foundation model that acts as a powerful perception backbone, and a lightweight, trainable action policy head.

\paragraph{Frozen Perception Backbone.} We utilize the complete, pre-trained Segment Anything Model 2 (SAM2) with all its parameters frozen. SAM2's role is to act as a powerful temporal-visual feature extractor. Given an initial prompt $p$ and a video stream $(I_0, \dots, I_T)$, its internal memory and tracking mechanisms allow it to identify and follow a specific object, producing a sequence of rich, object-aware feature representations $F_t$ at each timestep. These features, which encapsulate not only the object's appearance but also its temporal evolution, serve as the input to our action policy.

\paragraph{Trainable Action Policy Head.} For the action head, we adopt the powerful Action Chunking with Transformers (ACT) architecture \cite{act}. Instead of a simple MLP that predicts a single action, ACT is a Transformer-based policy that predicts a "chunk" or a sequence of future actions at each inference step. This approach is highly effective for generating smooth, temporally consistent trajectories.

The ACT policy head in our framework takes two inputs at each timestep $t$: the object-aware feature vector $F_t$ extracted by SAM2, and the robot's current proprioceptive state $Prop_t$ (i.e., its joint angles and gripper status). The output is a sequence of actions for a fixed future time horizon. We demonstrate the flexibility of our framework by implementing two common action representations:

\begin{itemize}
    \item Joint Space Control: The policy outputs a sequence of target joint angles for the robot arm, plus a binary value for the gripper state (open/closed).
    \item Task Space Control: The policy outputs a sequence of target 6D end-effector poses (position and orientation), plus the gripper state.
\end{itemize}

This combination of a powerful, frozen perception module and a state-of-the-art, trainable action policy forms the core of our SAM2Grasp architecture. By leveraging ACT, we ensure that our system can generate high-quality, smooth motor commands, while the SAM2 backbone provides the robust, promptable perception needed to resolve grasping ambiguity.

\subsection{Efficient Two-Stage Training Paradigm}

The architectural separation of a frozen backbone and a lightweight head unlocks a highly efficient two-stage training process.

\textbf{Stage 1: Offline Feature Extraction and Caching.}
Instead of costly end-to-end training, we first process our entire demonstration dataset $\mathcal{D}$ in an offline pass. For each demonstration, we provide the video frames $(I_0, \dots, I_T)$ and the corresponding initial prompt $p$ to the frozen SAM2 model. We then save the resulting sequence of intermediate features $(F_0, \dots, F_T)$ to disk. This step creates a new, pre-processed dataset $\mathcal{D}' = \{(F_t, a_t^*)\}$, where expert actions are paired directly with high-level, object-aware features.

\textbf{Stage 2: Action Head Training.}
With the pre-computed dataset $\mathcal{D}'$, training becomes a simple supervised learning problem. We train only the parameters $\theta$ of the action head $h_{\text{action}}$ to minimize a regression loss, such as the L2 norm (MSE), between the predicted action $a_t = h_{\text{action}}(F_t; \theta)$ and the ground-truth expert action $a_t^*$. The loss function is given by:

 \begin{equation}
\label{eq:classification}
\mathcal{L}(\theta) = \frac{1}{N} \sum_{t=1}^{N} \left\| h_{\text{action}}(F_t; \theta) - a_t^* \right\|_2^2
\end{equation}

This training paradigm provides significant advantages:

\begin{enumerate}
    \item \textbf{Speed}: Training a lightweight policy head (even a Transformer-based one like ACT) on pre-computed features is orders of magnitude faster than end-to-end training of the full visuomotor system.
    \item \textbf{Efficiency}: It dramatically lowers the hardware requirements (VRAM, compute), making the approach highly accessible.
    \item \textbf{Stability}: It preserves the powerful, general-purpose knowledge of the foundation model by keeping its weights frozen, preventing catastrophic forgetting or performance degradation due to overfitting on small robotics datasets.
\end{enumerate}

\subsection{Asynchronous Inference and Grasp Execution}

To achieve highly reactive and smooth control, we employ a sophisticated asynchronous execution strategy, inspired by the temporal ensembling technique proposed by ACT. In contrast to the original ACT implementation—which operates synchronously—our approach explicitly decouples the policy inference latency from the robot's real-time control loop by running two concurrent threads: a high-frequency Robot Control Thread and a lower-frequency Policy Inference Thread, communicating via a shared, time-indexed action queue.

\paragraph{Policy Inference Thread.}
Operating at a moderate rate (e.g., $f_p \approx 20$ Hz), this thread periodically acquires the latest camera observation $I_t$ and the robot's proprioceptive state $Prop_t$. Images are passed through the frozen SAM2 backbone to obtain the object-centric feature $F_t$, which, together with $Prop_t$, is fed into the ACT-based policy head. The policy outputs a chunk of predicted actions $A_t = [a_t, a_{t+1}, ..., a_{t+K}]$. These actions are enqueued into a shared buffer, with past-due entries continuously pruned to maintain only relevant, future or current actions.

\paragraph{Robot Control Thread.}
This thread runs at a high frequency (e.g., $f_c \approx 100$ Hz) to satisfy the robot's real-time control requirements. At each control tick, it queries the action buffer for all available predictions corresponding to the current time $t_{\text{now}}$. Due to the overlapping nature of action chunks generated by the inference thread, there may be multiple candidate actions for $t_{\text{now}}$, denoted as $\mathcal{A}_{t_{\text{now}}}$. The final control command $a_{\text{final}}$ is obtained by temporally ensembling these candidates. 
This ensembled action is then sent to the robot's low-level controller for immediate execution.

This asynchronous, temporal-ensemble strategy not only masks the latency of expensive vision-based policy inference, but also leverages the redundancy from overlapping action predictions to smooth out noise and improve robustness.

\section{EXPERIMENTS}

We conduct a comprehensive set of experiments in both simulated and real-world settings to validate the effectiveness, data efficiency, and robustness of SAM2Grasp. Our evaluation is designed to answer three key questions:

\begin{enumerate}
    \item How does our prompt-based approach compare against standard imitation learning methods and other forms of conditioning in a task with severe object-level multimodality?
    \item How robust is SAM2Grasp to significant visual occlusions compared to other methods?
\end{enumerate}

\subsection{Experimental Setup}

Our evaluation focuses on multi-object grasping tasks characterized by significant multi-modality and visual ambiguity.

\begin{figure}[htbp]
    \centering
    \includegraphics[width=0.8\columnwidth]{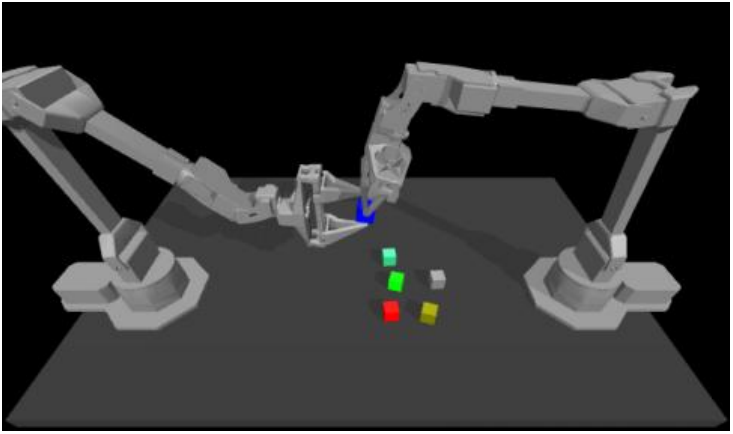}
    \caption{
        \textbf{Simulation Experiments.} 
    }
    \label{fig:simulation}
\end{figure}

\begin{figure}[htbp]
    \centering
    \includegraphics[width=0.9\columnwidth]{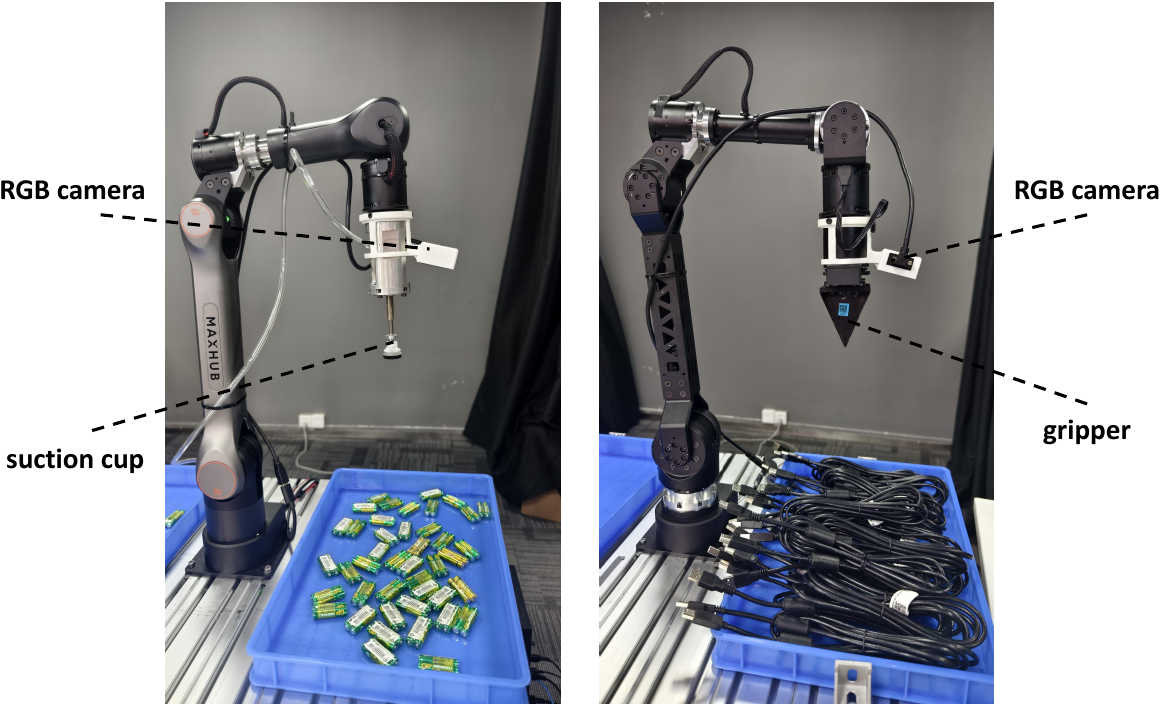}
    \caption{
        \textbf{Real-World Experiments.} 
    }
    \label{fig:real_experiment}
\end{figure}

\begin{itemize}
    \item Simulation Environment: As shown in Figure~\ref{fig:simulation}, We use a MuJoCo environment featuring a dual-arm setup for a challenging Multi-Object Pick and Handoff task. In each episode, 3 to 6 cubes of different colors are randomly placed on a table with random initial poses (position and orientation). The policy must grasp a cube with right robotic arm and hand it over to the left robotic arm. This task presents significant challenges in perception, multi-modality, and precise control.
    \item Real-World Environment: As shown in Figure~\ref{fig:real_experiment}, our real-world experiments are conducted on a 6-DoF MAXHUB A3 arm \cite{cvte} with a wrist-mounted camera(Intel RealSense D405). We use both a parallel gripper and a suction cup to demonstrate versatility. The task is Cluttered Bin Picking using different object types, including batteries(picked with the suction cup), power cables(grasped with the parallel gripper). A trial is successful if the arm picks up one object.
    \item Evaluation: All experiments are evaluated using Success Rate (SR), defined as the percentage of successful trials. For each main result, we report the mean SR over 400 trials in simulation / 200 trials in real-world.
\end{itemize}

\subsection{Compared Methods}

We compare SAM2Grasp against a carefully chosen set of baselines to isolate the contributions of different strategies for handling multimodality.

\begin{itemize}
    \item \textbf{ACT:} The standard deterministic Action Chunking Transformer \cite{act}. It receives only the raw RGB image and must resolve ambiguity on its own.
    \item \textbf{ACT-CVAE:} The original ACT model \cite{act}, which uses a CVAE to model a multi-modal action distribution. This represents the SOTA approach of tackling multimodality at the policy-output level.
    \item \textbf{ACT-CVAE-Condition (Strong Baseline):} To create a strong and fair baseline, we provide target information to the ACT-CVAE model directly at the input level. For each frame, we use a tracker to obtain the target object's bounding box and render it onto the RGB image. This method, which we term "visual conditioning," directly informs the policy which object to grasp.
    \item \textbf{SAM2Grasp (Ours):} Our proposed method, which uses a prompt to guide the frozen SAM2 backbone to extract object-aware features.
\end{itemize}

\subsection{Main Results}

\paragraph{Simulation.} As shown in Table \ref{tab:main_results}, The results lead to two key insights:

Conditioning at the Input is Superior to Modeling at the Output: Standard ACT (47.3\%) and ACT-CVAE (50.8\%) both struggle significantly. The minor improvement from CVAE suggests that simply modeling a multi-modal action distribution is insufficient to resolve the severe object-level ambiguity. In contrast, by providing a direct visual cue (ACT-CVAE-Condition), performance dramatically jumps to 81.0\%. This confirms our core hypothesis: explicitly conditioning the policy on the target identity is a far more effective strategy for object-level multimodality than trying to model the ambiguity at the output.

SAM2Grasp's Prompting is Superior to Simple Visual Conditioning: While visual conditioning is effective, SAM2Grasp (87.8\%) still demonstrates a significant performance gain over our strong baseline, ACT-CVAE-Condition (81.0\%). This performance gap reveals a key insight into \textit{how} the prompt is utilized. The ACT-CVAE-Condition baseline, with its visual cueing approach, still forces a standard vision backbone to learn the complex task of separating the cued object from a cluttered scene. In contrast, SAM2Grasp's prompt-driven perception leverages SAM2 to do this perceptual heavy-lifting upfront. The prompt actively guides the powerful, pre-trained SAM2 model to extract \textbf{clean, pre-filtered, object-centric features}, effectively giving the policy a much simpler and more focused problem to solve. This inherent advantage in perception directly translates to more precise and robust manipulation.

\begin{table}[htbp]
\centering
\caption{Simulation Results: Success Rate (\%) on the Multi--Object Pick-and-Handoff Task (400 trials)}
\label{tab:main_results}
\begin{tabular}{lcc}
\toprule
\textbf{Method} & \textbf{Simulation}\\ 
\midrule
ACT & $47.3 $  \\ 
ACT-CVAE & $50.8$ \\ 
ACT-CVAE-condition & $81.0$ \\ 
\textbf{SAM2Grasp (Ours)} & $\mathbf{87.8}$ \\ 
\bottomrule
\end{tabular}
\end{table}

\paragraph{Real World}

\begin{table}[htbp]
\centering
\caption{Real-World Results: Success Rate (\%) on the Cluttered Bin Picking Task (200 trials per category)}
\label{tab:real_world_results}
\begin{tabular*}{\columnwidth}{@{\extracolsep{\fill}} lccc @{}}
\toprule
\textbf{Method} & 
\textbf{\makecell{Batteries\\(Suction)}} & 
\textbf{\makecell{Cables\\(Grasping)}} & 
\textbf{Average} \\
\midrule
ACT & 33.0 & 50.0 & 41.5 \\
ACT-Condition & 72.9 & 70.5 & 71.7 \\
\textbf{SAM2Grasp (Ours)} & \textbf{98.5} & \textbf{95.5} & \textbf{97.0} \\
\bottomrule
\end{tabular*}
\end{table}

The standard ACT policy achieves a modest average success rate of 41.5\%. Its performance is inconsistent and insufficient for reliable deployment. This result highlights the challenge that even a powerful deterministic policy faces when confronted with object-level multimodality in a real-world setting.

Providing a direct visual cue (ACT-Condition) substantially boosts the average success rate to 71.7\%, once again confirming that resolving object-level ambiguity is critical. However, this visual cueing approach is still limited by its standard vision backbone.

Crucially, SAM2Grasp establishes a new level of performance and reliability, achieving an impressive 97.0\% average success rate (98.5\% for batteries, 95.5\% for cables). This substantial performance gap over even the strong ACT-Condition baseline powerfully demonstrates that the abstract, object-centric features extracted by SAM2's prompt-driven perception are highly robust to real-world challenges such as lighting variations, reflections, and diverse object geometries. This validates SAM2Grasp as a practical and reliable solution for real-world robotic manipulation

\begin{figure}[htbp]
    \centering
    \includegraphics[width=1.0\columnwidth]{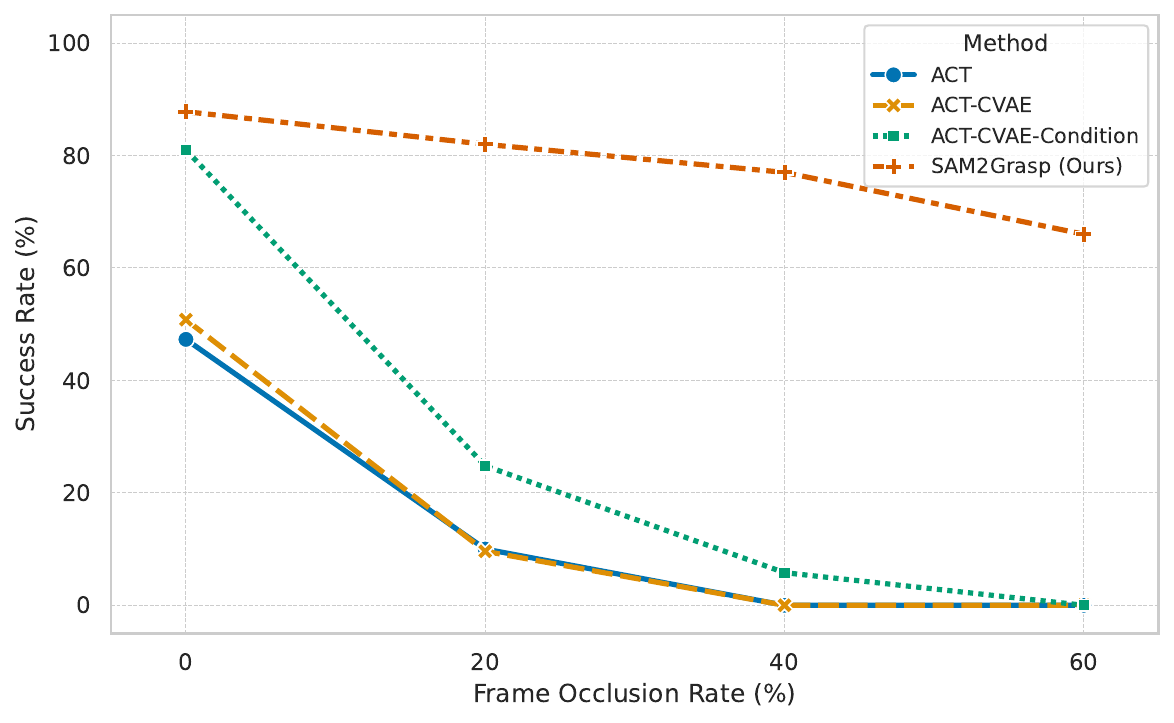}
    \caption{
        \textbf{Robustness to Visual Occlusion in Simulation.} 
        Success rate of all methods as a function of the frame occlusion rate ($p$). 
    SAM2Grasp exhibits remarkable resilience, showing only a graceful degradation in performance, whereas the baseline methods collapse under increasing visual perturbation. 
    This directly demonstrates the critical role of SAM2's built-in temporal memory for handling visual occlusion.
    }
    \label{fig:occlusion_plot}
\end{figure}

\begin{table}[htbp]
\centering
\caption{Robustness to Frame Occlusion in Simulation (Success Rate \%).}
\label{tab:occlusion_results}
\begin{tabular}{lcccc}
\toprule
\textbf{Method} & \textbf{p=0\%} & \textbf{p=20\%} & \textbf{p=40\%} & \textbf{p=60\%} \\ 
\midrule
ACT & 47.3 & 10.0 & 0.0 & 0.0 \\
ACT-CVAE & 50.8 & 9.6 & 0.0 & 0.0 \\
ACT-CVAE-Condition & 81.0 & 24.8 & 5.8 & 0.0 \\
\textbf{SAM2Grasp (Ours)} & \textbf{87.8} & \textbf{82.0} & \textbf{77.0} & \textbf{66.0} \\
\bottomrule
\end{tabular}
\end{table}

\subsection{Analysis of Robustness to Visual Occlusion}

To test the systems' resilience to real-world visual interruptions, we simulate occlusions by randomly blacking out a certain percentage ($p$) of frames in the input video stream during inference. 

The results, presented in Table \ref{tab:occlusion_results} and visualized in Figure \ref{fig:occlusion_plot}, are striking. As the occlusion rate increases, the performance of both baseline methods collapses; at a 40\% occlusion rate, their success rates plummet to near zero. This reveals their heavy reliance on a continuous stream of high-quality visual data.

In stark contrast, SAM2Grasp exhibits exceptional robustness. It maintains a 77\% success rate even with 40\% of frames missing, and still achieves a 66\% success rate under a severe 60\% occlusion rate. This remarkable resilience stems directly from our choice of foundation model. The primary reason is SAM2's built-in temporal memory. As a video-native model, SAM2 is explicitly designed to be robust to occlusions. When a frame is missing, its internal tracking mechanism propagates its belief about the object's state forward in time. This provides a continuous and stable feature stream to the policy head, effectively bridging the gaps in perception.

This experiment unequivocally demonstrates that leveraging a temporal-aware foundation model is a critical architectural choice for achieving robust robotic manipulation in imperfectly perceived environments.

\section{CONCLUSIONS}

We introduced SAM2Grasp, a novel framework that resolves the critical multi-modality problem in robotic grasping. By reformulating the task as a deterministic, prompt-conditioned problem and augmenting a frozen SAM2 foundation model with a lightweight action head, our method avoids the complexity of modeling multi-modal action distributions. Experiments in both simulation and the real world demonstrated that SAM2Grasp significantly outperforms state-of-the-art generative policies while being orders of magnitude more training-efficient.

 Future work can focus on integrating language models for automatic prompt generation from high-level commands and extending the framework to a broader range of manipulation skills beyond grasping.

\addtolength{\textheight}{-12cm}   





\end{document}